\documentclass[10pt]{article}
\usepackage[preprint]{tmlr}

\usepackage{amsmath,amsfonts,bm}

\def\eqref#1{equation~\ref{#1}}

\def\1{\bm{1}}

\DeclareMathAlphabet{\mathsfit}{\encodingdefault}{\sfdefault}{m}{sl}
\SetMathAlphabet{\mathsfit}{bold}{\encodingdefault}{\sfdefault}{bx}{n}

\usepackage{hyperref}
\usepackage{url}
\usepackage{graphicx}
\usepackage{booktabs}
\usepackage{longtable}
\usepackage{array}
\usepackage{calc}
\usepackage{etoolbox}
\makeatletter
\def\maxwidth{\ifdim\Gin@nat@width>\linewidth\linewidth\else\Gin@nat@width\fi}
\def\maxheight{\ifdim\Gin@nat@height>0.9\textheight0.9\textheight\else\Gin@nat@height\fi}
\makeatother
\setkeys{Gin}{width=\maxwidth,height=\maxheight,keepaspectratio}
\AtBeginEnvironment{longtable}{%
  \small
  \setlength{\tabcolsep}{3pt}%
}

\providecommand{\tightlist}{%
  \setlength{\itemsep}{0pt}\setlength{\parskip}{0pt}}

\title{Preservation Is Not Enough for Width Growth: Regime-Sensitive
Selection of Dense LM Warm Starts}

\author{\name Eren Unlu \\
  \addr Globeholder \\
  \addr Paris, France \\
  \addr ORCID: \href{https://orcid.org/0000-0001-5380-6305}{0000-0001-5380-6305}}

\def\month{04}
\def\year{2026}

\begin{document}

\maketitle

\begin{abstract}
Width expansion offers a practical route to reuse smaller
causal-language-model checkpoints, but selecting a widened warm start is
not solved by zero-step preservation alone. We study dense width growth
as a candidate-selection problem over full training states, including
copied weights, optimizer moments, and scheduler state. In a small-scale
TinyStories proxy, we compare exact-copy, perturbative,
asymmetric-reset, and structured non-clone warm starts under matched
continuation budgets. We evaluate zero-step preservation, short-lag
probe metrics, and downstream continuation utility in deterministic and
stochastic regimes. The picture is mixed and partially replicated
through a reduced-pool seed-1 check. Exact-copy symmetric warm starts
rank first in every completed 16-step probe and in the completed
stochastic 128-step continuations at seed-0 steps 1000 and 2000 plus
reduced seed-1 step 2000. By contrast, the structured non-clone
challenger wins deterministic 128-step continuation at seed-0 step 1000,
seed-0 step 2000, and seed-1 step 2000. Early escape from the inherited
cloned subspace is therefore not a universal selector: it helps in long
deterministic continuation, but it misleads at short lag and under
stochastic continuation. Across the completed studies, probe KL is the
most reliable low-cost selector overall, although the seed-0 step-1000
stochastic 128-step report shows that probe RMS can be the better
tie-breaker in some stochastic settings. The result is narrow but
useful: for dense width growth at this scale, preservation is not a
universal ranking criterion, and the best replacement signal depends on
both regime and lag budget.
\end{abstract}

\hypertarget{introduction}{%
\section{Introduction}\label{introduction}}

Training a family of causal language models from scratch is expensive
because each target size typically pays its own full optimization cost.
Progressive growth is attractive because it enables a different
workflow: train a smaller model first, then reuse that training state
when moving to a larger model \citep{ShenEtAl2022, YanoEtAl2025}. Recent
large-scale public evidence has made this workflow increasingly
credible, but much of the clearest dense-language-model evidence still
concentrates on progressive construction broadly or on depth growth more
than on dense width growth itself \citep{DuEtAl2024, YanoEtAl2025}.

Width growth remains attractive because it changes capacity without
altering sequence length or layer count, but it also creates a specific
initialization problem. Copy-based widening can preserve the parent
almost exactly at step 0, yet the resulting child may inherit strong
symmetries that delay or distort the use of newly added width. Recent
dense-width papers already show that this design space is real.
Function-preserving hidden-dimension expansion can accelerate dense
language-model pretraining, while width-progressive learning can fail if
signal preservation and symmetry breaking are not balanced carefully
\citep{SamraghEtAl2024, YuEtAl2026, MaEtAl2026}. Those papers, however,
do not settle a narrower operational question: once a lab has one fixed
parent checkpoint, which widened training state should be chosen next
under a small continuation budget?

That question is easy to underestimate because zero-step preservation
metrics are cheap and appealing. Parent-child KL, immediate validation
loss, and activation-statistics drift can all be measured without real
continuation. But selection work warns that initialization behavior can
be misleading, and that a little post-growth training can be more
predictive than initialization-only diagnostics \citep{KarpEtAl2024}.
For dense width growth, this concern is especially important because
copied weights, inherited optimizer moments, and scheduler state can all
affect whether the child remains trapped near the inherited cloned
subspace or begins to use genuinely new degrees of freedom
\citep{ShenEtAl2022, YuEtAl2026, MaEtAl2026}.

We therefore study dense width growth as a candidate-selection problem
over full training states. Our setting is intentionally narrow: a
decoder-only TinyStories proxy with fixed depth, tokenizer, and context
length, and a width-doubled child family evaluated under deterministic
and stochastic continuation regimes \citep{EldanLi2023}. Within this
setting we compare exact-copy, perturbative, asymmetric-reset, and
structured non-clone warm starts, then ask two questions. First, do
zero-step preservation metrics rank widened candidates well? Second, can
low-cost width-aware probe signals improve selection?

The answer depends on regime and horizon. Exact-copy symmetric warm
starts are the strongest default in every completed short 16-step probe
and every completed stochastic 128-step continuation. In deterministic
128-step continuation, however, a structured non-clone
reference-subspace warm start wins at seed-0 step 1000, seed-0 step
2000, and seed-1 step 2000. The implication is not that one new
warm-start recipe dominates. It is that preservation alone is not a
universal ranking criterion, while the most informative replacement
signal depends on both continuation regime and lag budget.

The paper makes four contributions.

\begin{itemize}
\tightlist
\item
  We reframe dense width growth as a selector problem over full training
  states rather than only a structural widening map
  \citep{ShenEtAl2022}.
\item
  We provide a single-GPU proxy study for this problem in a decoder-only
  TinyStories setting
  \citep{EldanLi2023, GeipingGoldstein2022, WortsmanEtAl2023}.
\item
  We show a mixed-regime result with reduced-pool partial replication:
  exact-copy dominates short-budget and stochastic continuation, while a
  structured non-clone challenger can win in longer deterministic
  continuation.
\item
  We show that selector quality is itself regime-sensitive: probe KL is
  the strongest overall low-cost default, while cloned-subspace escape
  becomes useful mainly in the deterministic long-horizon regime.
\end{itemize}

We do not claim a new general widening operator, a universal
width-growth recipe, or frontier-scale transfer. The contribution is a
dense-width selector study, together with a mechanism-oriented
interpretation, under constrained compute.

\hypertarget{related-work}{%
\section{Related Work}\label{related-work}}

\hypertarget{staged-growth-and-model-family-construction}{%
\subsection{Staged growth and model-family
construction}\label{staged-growth-and-model-family-construction}}

Staged-training work establishes the key framing used here: a growth
operator should act on the full training state, not only on weights
\citep{ShenEtAl2022}. That framing matters because optimizer moments and
scheduler state can alter post-growth behavior even when the widened
child is nearly loss-preserving at initialization. At the systems level,
recent progressive-training work also strengthens the practical
motivation for checkpoint reuse by showing that model families can be
constructed more economically through expansion \citep{YanoEtAl2025}. We
build on that framing, but focus on a narrower question than those
papers address: selection among dense width-expanded candidates from one
fixed parent checkpoint.

\hypertarget{dense-width-warm-starts-and-hyperparameter-aware-upscaling}{%
\subsection{Dense width warm starts and hyperparameter-aware
upscaling}\label{dense-width-warm-starts-and-hyperparameter-aware-upscaling}}

The closest dense-language-model overlap comes from hidden-dimension
expansion and warmstarting larger models from smaller ones. Scaling
Smart occupies the space of dense causal-LM width expansion with
function-preserving initialization \citep{SamraghEtAl2024}. Warmstarting
for Scaling Language Models and \(\mu\)pscaling further show that
warmstarts, perturbations, and hyperparameter transfer are already
active parts of the modern scaling pipeline
\citep{MallikEtAl2024, MaEtAl2026}. Two additional
transformer-language-model papers sharpen the novelty boundary.
\texttt{bert2BERT} extends reusable pretrained transformer growth
directly to BERT and GPT-style language models, while \texttt{LEMON}
studies lossless expansion together with the post-expansion scheduler
\citep{ChenEtAl2022Bert2Bert, WangEtAl2024LEMON}. Together, these papers
close off any novelty claim about parent-preserving dense widening
itself. The remaining question is narrower: given a fixed dense
width-expansion setting and a fixed small continuation budget, which
candidate training state should be selected?

\hypertarget{historical-function-preserving-widening-and-morphism}{%
\subsection{Historical function-preserving widening and
morphism}\label{historical-function-preserving-widening-and-morphism}}

The historical roots are older than current language-model work. Net2Net
made function-preserving widening a canonical acceleration strategy,
while Network Morphism generalized the idea to a broader family of
structure-preserving network changes
\citep{ChenGoodfellowShlens2015, WeiEtAl2016}. These papers matter as
claim boundaries. They establish that inheriting a parent function
inside a larger child is not the contribution of the present work. We
use that tradition as background for our candidate families, including
reference-subspace, but the contribution claimed here is diagnostic and
empirical rather than a new morphism operator.

\hypertarget{learned-transformer-growth-operators}{%
\subsection{Learned transformer growth
operators}\label{learned-transformer-growth-operators}}

Not all relevant prior work stays in hand-designed copy families. LiGO
learns linear width- and depth-growth operators for pretrained
transformers and reports strong results across language and vision
settings \citep{WangEtAl2023LiGO}. This matters because it already
occupies a large part of the ``transformer growth operator'' space. The
present paper therefore should not read as proposing a new operator
family or as opening the idea of learned transformer reuse. Its narrower
contribution is to ask how a lab should choose among
already-materialized dense width candidates when compute only permits
small post-growth budgets.

\hypertarget{width-specific-stability-and-symmetry-breaking}{%
\subsection{Width-specific stability and symmetry
breaking}\label{width-specific-stability-and-symmetry-breaking}}

Recent width-progressive work makes the central mechanism risk explicit.
SPARKLING argues that width expansion must balance signal preservation
with symmetry breaking, emphasizing RMS-scale consistency, asymmetric
optimizer-state reset, and rewarmup choices \citep{YuEtAl2026}. Recent
expansion work on inactive added neurons similarly warns that new
capacity can remain effectively unused after expansion
\citep{ChatzisEtAl2025}. These papers already occupy the claim that
width growth is not solved by naive copy or naive noise alone. Our study
differs by being dense decoder-only rather than primarily MoE-centered,
by treating the question as candidate selection rather than as recipe
design, and by evaluating whether the chosen candidate improves
downstream continuation utility under one fixed checkpoint and fixed
budget.

\hypertarget{growth-selection-and-early-dynamics}{%
\subsection{Growth selection and early
dynamics}\label{growth-selection-and-early-dynamics}}

LAG is the closest prior art on selection itself. Its core lesson is
that initialization behavior can mis-rank growth strategies, and that a
short period of real post-growth training can outperform
initialization-only diagnostics \citep{KarpEtAl2024}. We adopt that
warning directly. Where LAG is broad, our study is width-specific: we
ask whether dense width growth exposes a useful low-cost signal beyond
generic early loss. The present results do not support a universal
``escape beats preservation'' story, but they do support the narrower
claim that dense width selection is both regime-sensitive and
lag-budget-sensitive.

\hypertarget{proxy-study-methodology}{%
\subsection{Proxy-study methodology}\label{proxy-study-methodology}}

Because this project operates under tight single-GPU limits, the
legitimacy of proxy evidence matters. Cramming shows that
constrained-compute language-model experiments can still be
methodologically serious, while work on small-scale proxies for
large-scale transformer instabilities shows that some training phenomena
can be investigated credibly at reduced scale
\citep{GeipingGoldstein2022, WortsmanEtAl2023}. TinyStories is
especially useful in this setting because it supports controlled
decoder-only experiments where small models still exhibit nontrivial
generative behavior \citep{EldanLi2023}.

Taken together, the literature leaves a narrow but viable opening. Prior
work already covers function-preserving widening, reusable pretrained
transformer growth, learned growth operators, dense hidden-dimension
warm starts, width-stability mechanisms, and general growth selection.
What remains underexplored is the dense-width-specific question studied
here: how to choose among widened full training states from one fixed
parent checkpoint when compute only allows small post-growth budgets,
and how that answer changes across deterministic versus stochastic
continuation and across short versus longer lag budgets.

\hypertarget{problem-formulation-and-metrics}{%
\section{Problem Formulation and
Metrics}\label{problem-formulation-and-metrics}}

We define the parent checkpoint as a full training state

\[
P = \left(\theta^{(p)}, \omega^{(p)}, \sigma^{(p)}\right),
\]

where \(\theta^{(p)}\) denotes model parameters, \(\omega^{(p)}\)
denotes AdamW optimizer state, and \(\sigma^{(p)}\) denotes scheduler
state. A widened candidate is likewise a full training state

\[
c = \left(\theta_c, \omega_c, \sigma_c\right) \in \mathcal{C}(P),
\]

derived from the same parent checkpoint by one candidate-construction
recipe. This is the object selected in the paper.

The current repo uses two qualitatively different dense-width maps. Let
the width multiplier be \(m\). In clone-based exact copy, tensors on the
widened hidden dimension are repeated across the new clone axis, and
outgoing readout weights are scaled by factors such as \(1/m\) so the
child initially preserves the parent function under the cloned
representation. In \texttt{refsubspace\_statecopy}, the parent is
embedded into one reference slice, the non-reference slices are
initialized dormant at step 0, and the write-in versus read-out maps use
complementary \(\sqrt{m}\) and \(1/\sqrt{m}\) scalings so the parent
function is preserved while the child remains anchored to a single
inherited subspace rather than to symmetric clones. These are
implementation-level operator definitions rather than a general morphism
theory, and they match the widening maps in
\texttt{src/width\_growth/widen.py}.

For regime \(r \in \{\mathrm{det}, \mathrm{stoch}\}\) and continuation
horizon \(H\), let \(\ell_{r,c}(t_i)\) be the validation loss recorded
for candidate \(c\) at evaluation step \(t_i\), with
\(0=t_0 < t_1 < \cdots < t_n = H\). The downstream utility is
validation-loss area under the continuation curve,

\[
U_r^{(H)}(c)
=
\operatorname{AUC}_r^{(H)}(c)
=
\sum_{i=0}^{n-1}
\frac{\ell_{r,c}(t_i)+\ell_{r,c}(t_{i+1})}{2}
\left(t_{i+1}-t_i\right),
\]

where lower values are better. This matches the repo's trapezoidal
integration over the logged validation-loss trajectory.

Given a validation batch \(x\), the zero-step parent-child KL used in
the repo is

\[
\operatorname{KL}_0(c; x)
=
\frac{1}{BT}
\sum_{b=1}^{B}
\sum_{t=1}^{T}
\sum_{v}
p_p(v \mid x_{b,\le t})
\left[
\log p_p(v \mid x_{b,\le t})
-
\log p_c(v \mid x_{b,\le t})
\right].
\]

The zero-step mean RMS drift averages log-RMS mismatch across
hidden-state blocks:

\[
D_{\mathrm{RMS}}(c; x)
=
\frac{1}{L}
\sum_{\ell=1}^{L}
\left|
\log
\frac{\operatorname{RMS}(h_c^{(\ell)}(x))}
{\operatorname{RMS}(h_p^{(\ell)}(x))}
\right|.
\]

The escape metric is defined on the effective AdamW update rather than
on the raw gradient. For parameter tensor \(j\) at optimizer step \(t\),
the repo computes

\[
\Delta_j(c)
=
-
\frac{\eta}{1-\beta_1^t}
\frac{m_j}{\sqrt{v_j}/\sqrt{1-\beta_2^t}+\varepsilon}
-
\eta \lambda \theta_j,
\]

with learning rate \(\eta\), AdamW moments \((m_j, v_j)\), coefficients
\((\beta_1,\beta_2)\), stabilizer \(\varepsilon\), and weight decay
\(\lambda\). The metadata of each widened tensor defines a projector
\(\Pi_j\) onto the inherited low-dimensional subspace. The aggregate
escape score is

\[
E(c)
=
\frac{
\sum_j \left\| \Delta_j(c) - \Pi_j(\Delta_j(c)) \right\|_2^2
}{
\sum_j \left\| \Delta_j(c) \right\|_2^2
}.
\]

In clone-based exact-copy candidates, \(\Pi_j\) is a clone-mean
projector. In \texttt{refsubspace\_statecopy}, it is a keep-reference
projector. The selector regret reported by the paper is

\[
R_m(s)
=
U(s, \hat c_m(s))
-
\min_{c \in \mathcal{C}(s)} U(s, c),
\]

where \(m\) is a selector metric and \(\hat c_m(s)\) is the candidate
chosen by that metric for setting \(s\). Lower regret is better. When a
selector ties multiple candidates, the repo keeps best-tie and worst-tie
variants; the manuscript tables use best-tie regret unless stated
otherwise.

\hypertarget{experimental-setup}{%
\section{Experimental Setup}\label{experimental-setup}}

We study width growth as a selection problem over widened training
states. Each candidate consists of model parameters, optimizer state,
and scheduler state inherited or modified from the same parent
checkpoint \citep{ShenEtAl2022}. The canonical parent is a decoder-only
language model with 6 layers, hidden size 256, context length 256,
vocabulary size 8192, and 8,457,472 parameters. The widened child
doubles width to hidden size 512 while keeping depth, tokenizer, and
context length fixed. In our implementation, widening scales the number
of attention heads with the model width so attention head dimension
stays fixed, and the child output head is untied even when the parent
used tied embeddings because the exact-copy input-embedding and
output-projection expansions require different scaling.

The main study uses TinyStories as a dense decoder-only proxy benchmark
\citep{EldanLi2023}. In our pipeline, the canonical seed-0 parent uses
217,479 packed training sequences and 18,344 packed validation
sequences. The decoder uses learned token and position embeddings,
RMSNorm pre-normalized residual blocks, causal scaled-dot-product
attention, and SwiGLU MLPs with multiplier 4.0. We use this benchmark
because it is small enough for controlled single-GPU experimentation
while still supporting meaningful language-model behavior
\citep{EldanLi2023, GeipingGoldstein2022}.

The tokenizer is trained inside the repo rather than taken as a fixed
external artifact. It is a byte-level BPE tokenizer with NFKC
normalization, vocabulary 8192, minimum frequency 2, and a training
subset of 50,000 TinyStories examples. Each story is wrapped with
\texttt{\textless{}bos\textgreater{}} and
\texttt{\textless{}eos\textgreater{}}, flattened into a token stream,
and packed contiguously into 257-token rows so inputs and targets are
aligned 256-token next-token pairs. Parent optimization uses AdamW with
learning rate \texttt{3e-4}, minimum learning rate \texttt{3e-5}, betas
\texttt{(0.9,\ 0.95)}, weight decay \texttt{0.1}, and global gradient
clipping \texttt{1.0}. The scheduler is linear warmup for 100 steps
followed by cosine decay over 2000 total steps. Parent training uses
micro-batch size 12, gradient accumulation 4, and therefore an effective
batch of 48 packed sequences or 12,288 tokens per optimizer step.

The full seed-0 candidate pool contains \texttt{scratch\_large},
\texttt{exactcopy\_symmetric}, \texttt{exactcopy\_perturb\_symmetric},
\texttt{exactcopy\_asymreset}, \texttt{exactcopy\_asymreset\_rewarmup},
and \texttt{refsubspace\_statecopy}. The reduced seed-1 replication
keeps \texttt{scratch\_large}, \texttt{exactcopy\_symmetric},
\texttt{exactcopy\_perturb\_symmetric}, and
\texttt{refsubspace\_statecopy}. The last of these is the main
structured non-clone challenger: it embeds the parent exactly into one
active width slice, leaves the added slice dormant at step 0, and copies
optimizer state only into the preserved reference slice.

We evaluate candidates in two regimes. Deterministic continuation uses
fixed order, fixed seeds, and no dropout. Stochastic continuation uses
shuffled continuation plus a continuation-time dropout override
\texttt{p=0.1}. The canonical parent itself was trained with
\texttt{dropout=0.0}, so stochasticity is intentionally introduced after
growth rather than inherited from parent training. This makes the regime
split informative, but it also means the deterministic-versus-stochastic
comparison is a bundled contrast rather than a clean ablation of order
noise and dropout noise.

We study two post-growth budgets. Short lag uses 16 continuation steps
and acts as the paper's mini-LAG-style selector budget
\citep{KarpEtAl2024}. Long lag uses 128 continuation steps and serves as
the fixed-horizon utility target. For each candidate we log zero-step
validation loss, zero-step parent-child KL, zero-step mean RMS drift,
and zero-step cloned-subspace escape. During probe continuation we
aggregate \texttt{probe\_mean\_kl}, \texttt{probe\_mean\_rms\_drift},
and \texttt{probe\_mean\_escape}.

The primary downstream utility is validation-loss area under the
continuation curve, where lower AUC is better. We evaluate selectors by
candidate ranking, top-1 regret relative to the best candidate in the
same report, and rank correlation with downstream utility. This framing
keeps the paper focused on selection rather than on one final perplexity
number.

The current evidence base includes seed-0 short scans at steps 1000 and
2000 in deterministic and stochastic regimes, seed-0 deterministic
128-step continuation at steps 1000 and 2000, seed-0 stochastic 128-step
continuation at steps 1000 and 2000, and reduced seed-1 16-step and
128-step comparisons at step 2000. This evidence base supports the
narrow empirical claim set made below, but it does not justify
large-scale generalization claims, uncertainty estimation, or a clean
causal decomposition of the stochastic regime.

\hypertarget{protocol-tables}{%
\subsection{Protocol tables}\label{protocol-tables}}

The generated tables below keep the recipe definitions and completed
evidence matrix synchronized with the current artifact set.

\hypertarget{method-tables}{%
\section{Method Tables}\label{method-tables}}

This file is generated by
\texttt{python\ scripts/build\_rigor\_tables.py}.

\textbf{Table 3. Parent architecture, tokenization, and optimization
protocol.}

\begin{longtable}[]{@{}
  >{\raggedright\arraybackslash}p{(\columnwidth - 2\tabcolsep) * \real{0.50}}
  >{\raggedright\arraybackslash}p{(\columnwidth - 2\tabcolsep) * \real{0.50}}@{}}
\toprule
Component & Value \\
\midrule
\endhead
Dataset & \texttt{roneneldan/TinyStories} / \texttt{default} \\
Tokenizer & Byte-level BPE, NFKC normalization, vocab \texttt{8192}, min
frequency \texttt{2}, trained on \texttt{50000} examples \\
Packing & prepend \texttt{\textless{}bos\textgreater{}}, append
\texttt{\textless{}eos\textgreater{}}, flatten, then pack into
\texttt{257}-token rows for next-token prediction \\
Parent model & decoder-only LM with learned token and position
embeddings, \texttt{6} layers, \texttt{d\_model=256},
\texttt{n\_heads=8}, \texttt{mlp\_multiplier=4.0}, RMSNorm, SwiGLU \\
Parent dropout & \texttt{0.0} \\
Parent optimization & AdamW, \texttt{lr=0.0003}, \texttt{min\_lr=3e-05},
\texttt{betas=(0.9,\ 0.95)}, \texttt{weight\_decay=0.1},
\texttt{grad\_clip=1.0} \\
Parent scheduler & linear warmup \texttt{100} steps, cosine decay to
\texttt{min\_lr} over \texttt{2000} steps \\
Parent batch & micro-batch \texttt{12}, grad accumulation \texttt{4},
effective \texttt{48} sequences / \texttt{12288} tokens per optimizer
step \\
Parent evaluation & every \texttt{200} steps over \texttt{50} validation
batches \\
\bottomrule
\end{longtable}

\textbf{Table 4. Candidate recipes and state components.}

\begin{longtable}[]{@{}
  >{\raggedright\arraybackslash}p{(\columnwidth - 12\tabcolsep) * \real{0.13}}
  >{\raggedright\arraybackslash}p{(\columnwidth - 12\tabcolsep) * \real{0.13}}
  >{\raggedright\arraybackslash}p{(\columnwidth - 12\tabcolsep) * \real{0.13}}
  >{\raggedright\arraybackslash}p{(\columnwidth - 12\tabcolsep) * \real{0.13}}
  >{\raggedleft\arraybackslash}p{(\columnwidth - 12\tabcolsep) * \real{0.17}}
  >{\raggedleft\arraybackslash}p{(\columnwidth - 12\tabcolsep) * \real{0.17}}
  >{\raggedright\arraybackslash}p{(\columnwidth - 12\tabcolsep) * \real{0.13}}@{}}
\toprule
Recipe & Init & Opt state & Sched & Perturb & Warmup & Role \\
\midrule
\endhead
scratch-large & scratch & fresh & constant & 0.0000 & 0 & Fresh large
child with no inherited weights or optimizer state. \\
exact-copy & exact-copy & symmetric & constant & 0.0000 & 0 & Exact
function-preserving width copy with symmetric AdamW state
inheritance. \\
exact-copy + perturb & exact-copy & symmetric & constant & 0.0010 & 0 &
Exact-copy widened child with small isotropic post-copy perturbation. \\
exact-copy + asym-reset & exact-copy & asym-reset & constant & 0.0000 &
0 & Exact-copy widened child with new-clone AdamW moments reset out of
the cloned subspace. \\
exact-copy + asym-reset + rewarm & exact-copy & asym-reset & fresh
cosine & 0.0000 & 2 & Asymmetric-reset widened child probed under a
fresh global cosine rewarmup. \\
ref-subspace & ref-subspace & ref-slice copy & constant & 0.0000 & 0 &
Embed the parent exactly into one active width slice, keep the added
slice dormant at step 0, and copy optimizer state only into the
preserved reference slice. \\
\bottomrule
\end{longtable}

\textbf{Table 5. Completed evidence matrix.}

\begin{longtable}[]{@{}
  >{\raggedright\arraybackslash}p{(\columnwidth - 12\tabcolsep) * \real{0.12}}
  >{\raggedright\arraybackslash}p{(\columnwidth - 12\tabcolsep) * \real{0.12}}
  >{\raggedleft\arraybackslash}p{(\columnwidth - 12\tabcolsep) * \real{0.17}}
  >{\raggedright\arraybackslash}p{(\columnwidth - 12\tabcolsep) * \real{0.12}}
  >{\raggedleft\arraybackslash}p{(\columnwidth - 12\tabcolsep) * \real{0.17}}
  >{\raggedright\arraybackslash}p{(\columnwidth - 12\tabcolsep) * \real{0.12}}
  >{\raggedleft\arraybackslash}p{(\columnwidth - 12\tabcolsep) * \real{0.17}}@{}}
\toprule
Setting & Regime & Horizon & Winner & Delta AUC & Best selector & Best
regret \\
\midrule
\endhead
S0 / 1000 / 16D & deterministic & 16 & exact-copy & 0.0791 & zero loss,
zero KL, probe KL, probe RMS & 0.0000 \\
S0 / 1000 / 16S & stochastic & 16 & exact-copy & 0.1093 & zero loss,
zero KL, probe RMS & 0.0000 \\
S0 / 1000 / 128D & deterministic & 128 & ref-subspace & -3.9633 & probe
KL, probe escape & 0.0000 \\
S0 / 1000 / 128S & stochastic & 128 & exact-copy & 0.1047 & zero loss,
zero KL, probe RMS & 0.0000 \\
S0 / 2000 / 16D & deterministic & 16 & exact-copy & 0.0324 & zero loss,
zero KL, probe KL, probe RMS & 0.0000 \\
S0 / 2000 / 16S & stochastic & 16 & exact-copy & 0.0892 & zero loss,
zero KL, probe KL, probe RMS & 0.0000 \\
S0 / 2000 / 128D & deterministic & 128 & ref-subspace & -0.6090 & probe
KL, probe escape & 0.0000 \\
S0 / 2000 / 128S & stochastic & 128 & exact-copy & 0.8816 & zero loss,
zero KL, probe KL, probe RMS & 0.0000 \\
S1 / 2000 / 16D & deterministic & 16 & exact-copy & 0.0483 & probe KL,
probe RMS & 0.0000 \\
S1 / 2000 / 16S & stochastic & 16 & exact-copy & 0.0864 & probe KL,
probe RMS & 0.0000 \\
S1 / 2000 / 128D & deterministic & 128 & ref-subspace & -0.5574 & zero
loss, zero KL, probe KL, probe escape & 0.0000 \\
S1 / 2000 / 128S & stochastic & 128 & exact-copy & 0.9876 & probe KL,
probe RMS & 0.0000 \\
\bottomrule
\end{longtable}

\hypertarget{results}{%
\section{Results}\label{results}}

The main empirical pattern depends on both regime and horizon, with a
reduced-pool seed-1 replication rather than a universal selector claim.

\hypertarget{short-lag-continuation-favors-exact-copy-warm-starts}{%
\subsection{Short-lag continuation favors exact-copy warm
starts}\label{short-lag-continuation-favors-exact-copy-warm-starts}}

At short lag, exact-copy remains the strongest default. In the seed-0
full pool, every completed 16-step probe ranks
\texttt{exactcopy\_symmetric} first. The same pattern survives in the
reduced seed-1 study: the deterministic 16-step report ranks
\texttt{exactcopy\_symmetric} first with AUC 40.0196, ahead of
\texttt{refsubspace\_statecopy} at 40.0680, and the stochastic 16-step
report also ranks \texttt{exactcopy\_symmetric} first with AUC 40.0591,
ahead of \texttt{refsubspace\_statecopy} at 40.1455. These reduced
16-step runs are the closest thing in the current study to a mini-LAG
baseline. They already outperform zero-step preservation metrics as a
selector, but the lag budget is still too short to reveal the later
deterministic reversal \citep{KarpEtAl2024}.

\hypertarget{deterministic-long-horizon-reveals-a-structured-non-clone-advantage}{%
\subsection{Deterministic long horizon reveals a structured non-clone
advantage}\label{deterministic-long-horizon-reveals-a-structured-non-clone-advantage}}

At longer deterministic horizon, the ranking flips. In the seed-0
step-1000 deterministic 128-step continuation,
\texttt{refsubspace\_statecopy} beats \texttt{exactcopy\_symmetric} by
3.9633 AUC. In the seed-0 step-2000 deterministic 128-step continuation,
it wins again by 0.6090 AUC. The reduced seed-1 deterministic
replication preserves the same direction, with
\texttt{refsubspace\_statecopy} beating \texttt{exactcopy\_symmetric} by
0.5574 AUC. The point is not that \texttt{reference-subspace} is a
universally better recipe. The point is that deterministic continuation
with a long enough horizon can reward a structured non-clone warm start
that initially preserves the parent almost exactly but does not force
the child to remain inside the cloned-copy symmetry pattern. Table 1
reports the corresponding per-step normalization: the seed-0 step-1000
deterministic reversal is large at about \texttt{-0.0310} mean loss per
step, while the step-2000 deterministic reversals are materially smaller
at about \texttt{-0.0048} and \texttt{-0.0044} per step.

The full seed-0 pool no longer needs to remain implicit. The
supplementary candidate tables show that the deterministic long-horizon
reversal is not explained away by the asymmetric-reset families. In both
seed-0 deterministic 128-step settings, \texttt{refsubspace\_statecopy}
still ranks above \texttt{exactcopy\_asymreset} and
\texttt{exactcopy\_asymreset\_rewarmup}, while \texttt{scratch\_large}
remains far behind the seeded warm starts.

\hypertarget{stochastic-long-horizon-restores-the-exact-copy-default}{%
\subsection{Stochastic long horizon restores the exact-copy
default}\label{stochastic-long-horizon-restores-the-exact-copy-default}}

Under stochastic continuation, the deterministic reversal disappears. In
the seed-0 step-1000 stochastic 128-step continuation,
\texttt{exactcopy\_symmetric} stays ahead of
\texttt{refsubspace\_statecopy} by 0.1047 AUC. In the seed-0 step-2000
stochastic 128-step continuation, it stays ahead again by 0.8816 AUC.
The reduced seed-1 stochastic replication keeps the same ordering, with
\texttt{exactcopy\_symmetric} ahead by 0.9876 AUC. Taken together with
the short-lag results, this rules out a simple ``escape beats
preservation'' story. The current evidence instead says that structured
non-clone escape helps only in a specific corner of the space:
low-stochasticity continuation with enough horizon for the extra
capacity to become useful. Even there, the regime contrast should be
read carefully because the stochastic regime bundles shuffled
continuation with a continuation-time dropout override.

\hypertarget{selector-quality-is-regime--and-budget-sensitive}{%
\subsection{Selector quality is regime- and
budget-sensitive}\label{selector-quality-is-regime--and-budget-sensitive}}

No single low-cost selector is correct in every tested setting.
\texttt{probe\_mean\_kl} is still the strongest low-cost selector
overall, but it is no longer exact in every seed-0 long-horizon report:
in the seed-0 step-1000 stochastic 128-step run it selects
\texttt{refsubspace\_statecopy} and incurs 0.1047 regret, while
\texttt{probe\_mean\_rms\_drift} picks \texttt{exactcopy\_symmetric}
with zero regret. Even so, \texttt{probe\_mean\_kl} has zero regret in
the seed-0 deterministic 128-step runs, the seed-0 step-2000 stochastic
128-step run, and every reduced seed-1 setting currently exported in the
study tables. By contrast, \texttt{probe\_mean\_escape} becomes useful
only in the deterministic long-horizon regime. In the reduced seed-1
study it selects \texttt{refsubspace\_statecopy} in both short and long
deterministic runs, but only the long deterministic run makes that
choice correct; in the stochastic 128-step run it incurs 0.9876 regret.

The selector result has two parts. First, zero-step preservation is too
weak because it often ties together materially different continuation
behaviors \citep{KarpEtAl2024, ShenEtAl2022}. The main selector table
now makes that comparison explicit by adding
\texttt{zero\_step\_val\_loss} alongside \texttt{zero\_step\_kl}, and
neither zero-step baseline displaces \texttt{probe\_mean\_kl} as the
strongest overall low-cost selector. Second, the replacement signal is
not one universal escape diagnostic; it is a regime-sensitive mixture in
which probe KL is the most reliable default and escape is best treated
as a mechanism variable that becomes informative only in long
deterministic continuation.

The direct mini-LAG baseline sharpens this point further. If the
matching 16-step report is used directly to choose a candidate for the
corresponding 128-step target by short-run AUC, that selector is exact
in all three stochastic long-horizon settings but misses all three
deterministic long-horizon reversals. The companion short-run final-loss
baseline is more mixed: it recovers the seed-0 step-1000 deterministic
reversal, but it misses the seed-0 step-1000 stochastic winner and both
step-2000 deterministic reversals. The rank correlation of the 16-step
AUC selector with 128-step utility remains high. That makes the failure
informative rather than dismissive: a little lag helps, but a short lag
is still not sufficient when the deterministic ranking flips late.

\hypertarget{figure-supported-synthesis}{%
\subsection{Figure-supported
synthesis}\label{figure-supported-synthesis}}

Figure 1 summarizes the main paper result. It plots the AUC gap
\texttt{refsubspace\ -\ exactcopy}, so negative values favor
\texttt{refsubspace\_statecopy} and positive values favor
\texttt{exactcopy\_symmetric}. In the deterministic panel, every
available series moves downward from the 16-step probe to the 128-step
continuation, and the seed-0 step-1000 series plus both step-2000 series
cross below zero. In the stochastic panel, all available series remain
positive: the seed-0 step-1000 series stays close to zero at 128 steps,
while the seed-0 and seed-1 step-2000 series move farther positive. This
is the cleanest compact statement of the evidence: deterministic
long-horizon continuation can reverse the short-lag ranking, while
stochastic continuation does not. Table 1 reports the corresponding
per-step normalization so the reader does not need to infer effect size
only from raw AUC, which scales with horizon length.

\begin{figure}
\centering
\includegraphics{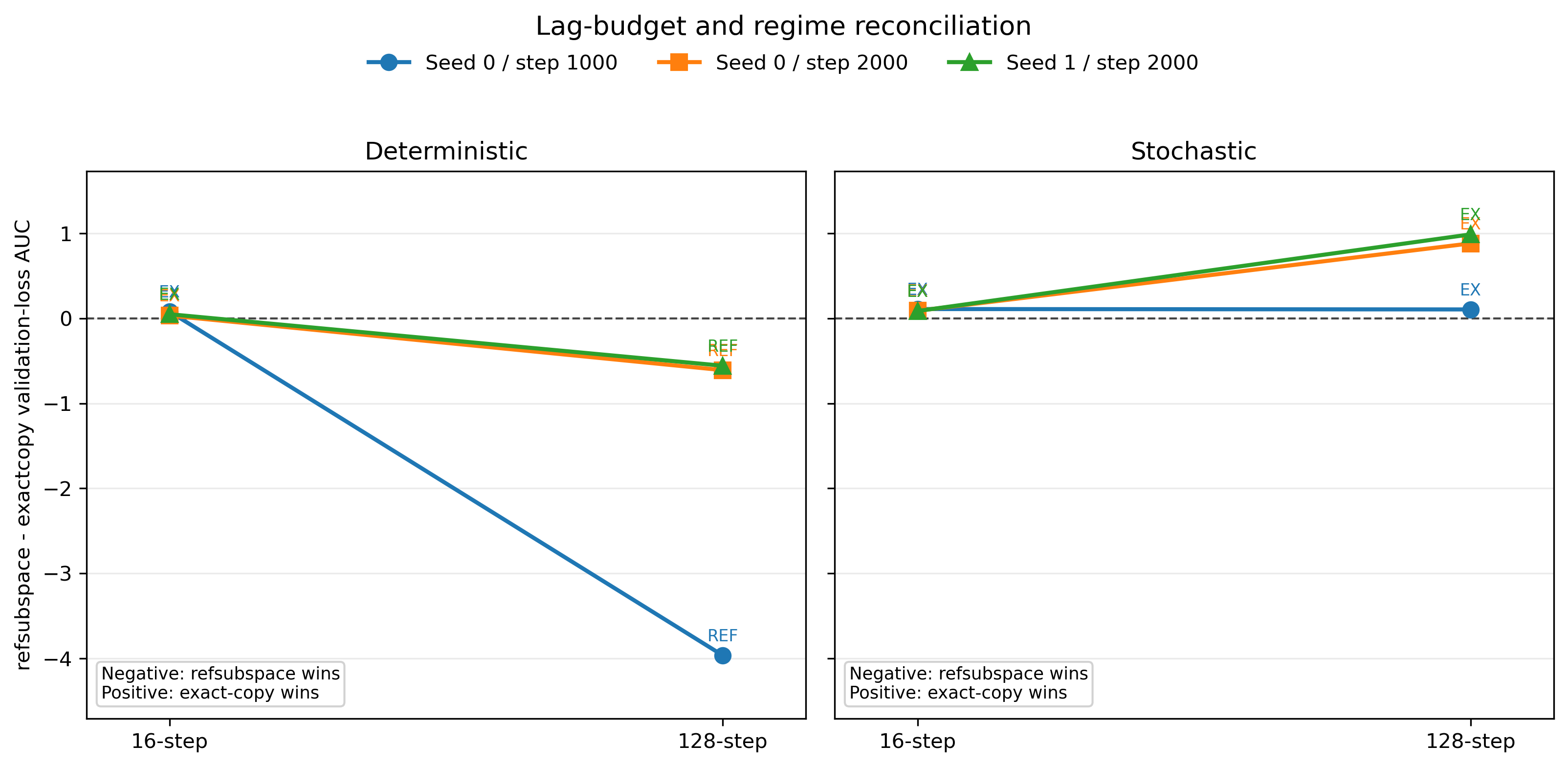}
\caption{Lag-budget and regime reconciliation for the main structured
non-clone challenger. Each point shows the validation-loss AUC gap
between refsubspace\_statecopy and exactcopy\_symmetric, computed as
refsubspace minus exactcopy, so negative values favor
refsubspace\_statecopy and positive values favor exactcopy\_symmetric.
In the deterministic panel, every available series moves downward from
the 16-step probe to the 128-step continuation, and the seed-0 step-1000
series plus both step-2000 series cross below zero. In the stochastic
panel, all available series remain positive: the seed-0 step-1000 series
stays close to zero at 128 steps, while the seed-0 and seed-1 step-2000
series move farther positive.}
\end{figure}

\hypertarget{selector-figure-synthesis}{%
\subsection{Selector figure synthesis}\label{selector-figure-synthesis}}

Figure 2 summarizes the selector-quality result directly. It shows
selector top-1 regret across completed settings for
\texttt{Zero-step\ loss}, \texttt{Zero-step\ KL}, \texttt{Probe\ KL},
\texttt{Probe\ RMS}, and \texttt{Probe\ Escape}. The left panel covers
the full seed-0 study; the right panel covers the reduced seed-1
replication. \texttt{Probe\ KL} is the strongest overall low-cost
selector, staying at zero regret across all reduced seed-1 settings and
all seed-0 deterministic long-horizon settings, but the seed-0 step-1000
stochastic 128-step run shows a small \texttt{Probe\ RMS} win over
\texttt{Probe\ KL}. \texttt{Probe\ escape} reaches zero regret only in
deterministic 128-step continuation, exactly where the structured
non-clone challenger actually wins. \texttt{Probe\ RMS} remains
competitive in short and stochastic settings but misses both
deterministic long-horizon reversals. This is the most compact visual
argument that the paper should not be framed as ``escape always wins'';
the correct claim is that selector quality itself is regime- and
lag-budget-sensitive.

\begin{figure}
\centering
\includegraphics{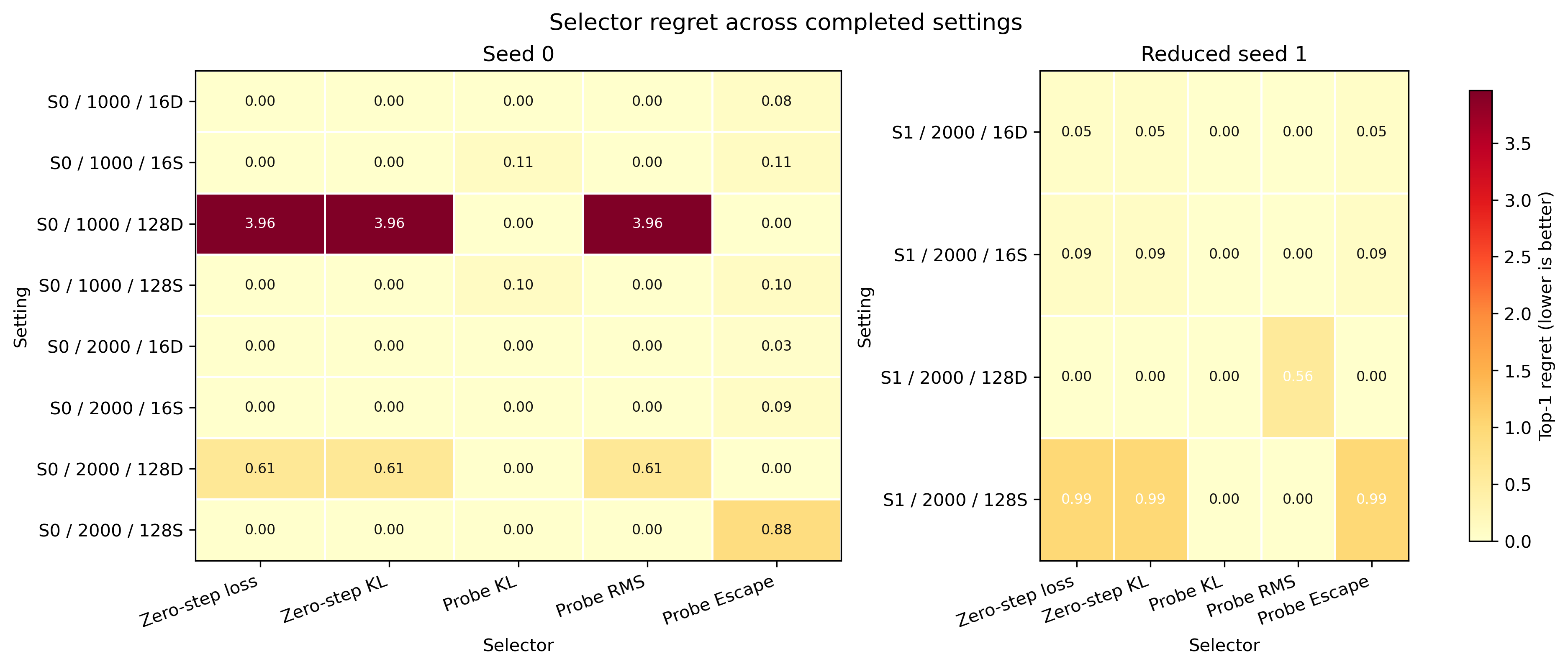}
\caption{Selector top-1 regret across completed settings. Each cell
shows the regret of a selector relative to the best candidate in the
same report, with lower values better. The left panel covers the full
seed-0 study; the right panel covers the reduced seed-1 replication.
Probe KL is the strongest overall low-cost selector, staying at zero
regret across all reduced seed-1 settings and all seed-0 deterministic
long-horizon settings, but the seed-0 step-1000 stochastic 128-step run
shows a small Probe RMS win over Probe KL. Probe escape reaches zero
regret only in deterministic 128-step continuation, where the structured
non-clone challenger actually wins. Probe RMS remains competitive in
short and stochastic settings but misses both deterministic long-horizon
reversals. Zero-step KL is often acceptable at short lag, but it fails
when deterministic long-horizon continuation favors
refsubspace\_statecopy.}
\end{figure}

\hypertarget{compact-numerical-tables}{%
\subsection{Compact numerical tables}\label{compact-numerical-tables}}

The figures expose the two main qualitative reversals, but the numeric
tables still matter. Table 1 keeps only the comparison that matters for
the main challenger family: \texttt{refsubspace\ -\ exactcopy}. All
completed 16-step settings stay positive, both deterministic seed-0
128-step settings are negative, the reduced seed-1 deterministic
128-step setting is negative, and all available stochastic 128-step
settings remain positive. Table 2 does the same for selector quality.
\texttt{Probe\ KL} is the strongest low-cost selector overall, but it is
not exact in every seed-0 long-horizon case after the new step-1000
stochastic report; \texttt{Probe\ Escape} is exact only in deterministic
128-step continuation.

\begingroup
\small
\setlength{\tabcolsep}{4pt}

\textbf{Table 1. Compact regime summary.}

\begin{longtable}[]{@{}
  >{\raggedright\arraybackslash}p{(\columnwidth - 10\tabcolsep) * \real{0.14}}
  >{\raggedright\arraybackslash}p{(\columnwidth - 10\tabcolsep) * \real{0.14}}
  >{\raggedleft\arraybackslash}p{(\columnwidth - 10\tabcolsep) * \real{0.18}}
  >{\raggedleft\arraybackslash}p{(\columnwidth - 10\tabcolsep) * \real{0.18}}
  >{\raggedleft\arraybackslash}p{(\columnwidth - 10\tabcolsep) * \real{0.18}}
  >{\raggedleft\arraybackslash}p{(\columnwidth - 10\tabcolsep) * \real{0.18}}@{}}
\toprule
Setting & Winner & Exact-copy AUC & Ref-subspace AUC & Delta AUC
(\texttt{ref\ -\ exact}) & Mean delta / step \\
\midrule
\endhead
S0 / 1000 / 16D & exact-copy & 44.5945 & 44.6737 & 0.0791 & 0.0049 \\
S0 / 1000 / 16S & exact-copy & 44.1903 & 44.2996 & 0.1093 & 0.0068 \\
S0 / 1000 / 128D & ref-subspace & 360.9172 & 356.9538 & -3.9633 &
-0.0310 \\
S0 / 1000 / 128S & exact-copy & 351.1863 & 351.2910 & 0.1047 & 0.0008 \\
S0 / 2000 / 16D & exact-copy & 39.9548 & 39.9872 & 0.0324 & 0.0020 \\
S0 / 2000 / 16S & exact-copy & 39.9700 & 40.0591 & 0.0892 & 0.0056 \\
S0 / 2000 / 128D & ref-subspace & 320.9953 & 320.3863 & -0.6090 &
-0.0048 \\
S0 / 2000 / 128S & exact-copy & 320.0455 & 320.9271 & 0.8816 & 0.0069 \\
S1 / 2000 / 16D & exact-copy & 40.0196 & 40.0680 & 0.0483 & 0.0030 \\
S1 / 2000 / 16S & exact-copy & 40.0591 & 40.1455 & 0.0864 & 0.0054 \\
S1 / 2000 / 128D & ref-subspace & 321.6701 & 321.1126 & -0.5574 &
-0.0044 \\
S1 / 2000 / 128S & exact-copy & 320.7028 & 321.6904 & 0.9876 & 0.0077 \\
\bottomrule
\end{longtable}

\textbf{Table 2. Selector top-1 regret.}

\begin{longtable}[]{@{}lrrrrr@{}}
\toprule
Setting & Zero-step loss & Zero-step KL & Probe KL & Probe RMS & Probe
Escape \\
\midrule
\endhead
S0 / 1000 / 16D & 0.0000 & 0.0000 & 0.0000 & 0.0000 & 0.0791 \\
S0 / 1000 / 16S & 0.0000 & 0.0000 & 0.1093 & 0.0000 & 0.1093 \\
S0 / 1000 / 128D & 3.9633 & 3.9633 & 0.0000 & 3.9633 & 0.0000 \\
S0 / 1000 / 128S & 0.0000 & 0.0000 & 0.1047 & 0.0000 & 0.1047 \\
S0 / 2000 / 16D & 0.0000 & 0.0000 & 0.0000 & 0.0000 & 0.0324 \\
S0 / 2000 / 16S & 0.0000 & 0.0000 & 0.0000 & 0.0000 & 0.0892 \\
S0 / 2000 / 128D & 0.6090 & 0.6090 & 0.0000 & 0.6090 & 0.0000 \\
S0 / 2000 / 128S & 0.0000 & 0.0000 & 0.0000 & 0.0000 & 0.8816 \\
S1 / 2000 / 16D & 0.0483 & 0.0483 & 0.0000 & 0.0000 & 0.0483 \\
S1 / 2000 / 16S & 0.0864 & 0.0864 & 0.0000 & 0.0000 & 0.0864 \\
S1 / 2000 / 128D & 0.0000 & 0.0000 & 0.0000 & 0.5574 & 0.0000 \\
S1 / 2000 / 128S & 0.9876 & 0.9876 & 0.0000 & 0.0000 & 0.9876 \\
\bottomrule
\end{longtable}

\textbf{Table 3. Direct \texttt{16}-step mini-LAG baselines for
\texttt{128}-step targets.}

\begin{longtable}[]{@{}llrlrr@{}}
\toprule
Target & \texttt{16}-step AUC pick & Regret @ \texttt{128} &
\texttt{16}-step final pick & Regret @ \texttt{128} & Spearman \\
\midrule
\endhead
S0 / 1000 / 128D & exact-copy & 3.9633 & ref-subspace & 0.0000 &
0.8286 \\
S0 / 1000 / 128S & exact-copy & 0.0000 & exact-copy + perturb & 0.1539 &
0.9429 \\
S0 / 2000 / 128D & exact-copy & 0.6090 & exact-copy & 0.6090 & 0.9429 \\
S0 / 2000 / 128S & exact-copy & 0.0000 & exact-copy & 0.0000 & 1.0000 \\
S1 / 2000 / 128D & exact-copy & 0.5574 & exact-copy & 0.5574 & 0.8000 \\
S1 / 2000 / 128S & exact-copy & 0.0000 & exact-copy & 0.0000 & 1.0000 \\
\bottomrule
\end{longtable}

\textbf{Interpretation.}

Lower AUC is better. Negative \texttt{ref\ -\ exact} means
\texttt{ref-subspace} beats \texttt{exact-copy}. The per-step column
normalizes the same gap by horizon length so short and long settings can
be compared without raw-AUC scaling alone. Lower selector regret is
better. Zero means the selector picked a winner up to ties. Table 3
treats the corresponding \texttt{16}-step run as a direct mini-LAG
baseline for the matching \texttt{128}-step target.

\endgroup

\hypertarget{claim-audit-tables}{%
\subsection{Claim audit tables}\label{claim-audit-tables}}

The tables below act as an internal red-team check on the paper's
central statements. They are generated from the current artifact exports
rather than maintained by hand.

\hypertarget{claim-audit-tables-1}{%
\section{Claim Audit Tables}\label{claim-audit-tables-1}}

This file is generated by
\texttt{python\ scripts/build\_rigor\_tables.py}.

\textbf{Table 6. Aggregate selector statistics across all completed
settings.}

\begin{longtable}[]{@{}
  >{\raggedright\arraybackslash}p{(\columnwidth - 8\tabcolsep) * \real{0.16}}
  >{\raggedleft\arraybackslash}p{(\columnwidth - 8\tabcolsep) * \real{0.21}}
  >{\raggedleft\arraybackslash}p{(\columnwidth - 8\tabcolsep) * \real{0.21}}
  >{\raggedleft\arraybackslash}p{(\columnwidth - 8\tabcolsep) * \real{0.21}}
  >{\raggedleft\arraybackslash}p{(\columnwidth - 8\tabcolsep) * \real{0.21}}@{}}
\toprule
Selector & Mean regret & Max regret & Zero-regret settings & Mean
Spearman to utility \\
\midrule
\endhead
Zero-step loss & 0.4745 & 3.9633 & 7 & 0.4191 \\
Zero-step KL & 0.4745 & 3.9633 & 7 & 0.4191 \\
Probe KL & 0.0178 & 0.1093 & 10 & 0.9619 \\
Probe RMS & 0.4275 & 3.9633 & 9 & 0.8571 \\
Probe Escape & 0.2015 & 0.9876 & 3 & -0.5000 \\
\bottomrule
\end{longtable}

\textbf{Table 7. Internal claim audit against current evidence.}

\begin{longtable}[]{@{}
  >{\raggedright\arraybackslash}p{(\columnwidth - 6\tabcolsep) * \real{0.25}}
  >{\raggedright\arraybackslash}p{(\columnwidth - 6\tabcolsep) * \real{0.25}}
  >{\raggedright\arraybackslash}p{(\columnwidth - 6\tabcolsep) * \real{0.25}}
  >{\raggedright\arraybackslash}p{(\columnwidth - 6\tabcolsep) * \real{0.25}}@{}}
\toprule
Claim ID & Audited statement & Status & Quantitative support \\
\midrule
\endhead
C1 & Exact-copy symmetric wins every completed 16-step setting. & pass &
6/6 short settings \\
C2 & Reference-subspace wins every completed deterministic 128-step
setting. & pass & 3/3 deterministic long settings \\
C3 & Exact-copy symmetric wins every completed stochastic 128-step
setting. & pass & 3/3 stochastic long settings \\
C4 & Probe KL is the strongest overall low-cost selector by mean top-1
regret. & pass & Zero-step loss=0.4745, Zero-step KL=0.4745, Probe
KL=0.0178, Probe RMS=0.4275, Probe Escape=0.2015 \\
C5 & Probe escape is exact only in deterministic 128-step settings. &
pass & zero-regret escape: 3/3 deterministic long, 0/9 all other
settings \\
\bottomrule
\end{longtable}

\hypertarget{discussion}{%
\section{Discussion}\label{discussion}}

The completed evidence does not support a universal rule such as
``escape beats preservation'' or ``exact copy is always enough.'' A
stricter statement fits the data better. Exact-copy symmetric warm
starts are the best default when the post-growth budget is short and
when continuation contains meaningful stochasticity. A structured
non-clone challenger becomes competitive, and in the current study wins,
only when continuation is deterministic and long enough for the added
width to matter. It is a selector result, not a recipe result.

That distinction matters because widening papers are easy to read as
method papers. The present study does not justify that style of claim.
The stronger conclusion is simpler: widened candidates should be
evaluated as full training states, and their ranking can change
materially with regime and lag budget even when several candidates are
almost indistinguishable under zero-step preservation metrics
\citep{ShenEtAl2022}.

For practitioners operating under a small post-growth budget, the safest
current default is still \texttt{exactcopy\_symmetric}. It wins every
completed short 16-step comparison and every completed stochastic
long-horizon comparison in this study. If the target workflow is
relatively deterministic and can afford a longer post-growth screen,
then a structured non-clone challenger deserves explicit evaluation
rather than dismissal based on zero-step preservation alone.

The low-cost selector implication is similarly narrow.
\texttt{probe\_mean\_kl} is the strongest overall default selector in
the completed studies even after adding \texttt{zero\_step\_val\_loss}
as an explicit baseline. Cloned-subspace escape is better understood as
a mechanism variable than as a universal ranking rule. It becomes
informative when deterministic continuation exposes clone trapping, but
it misleads in stochastic or too-short settings.

The regime split is central to the paper. Recent width-growth papers
make symmetry-breaking explanations plausible in general
\citep{YuEtAl2026, MaEtAl2026}. In our study, a reasonable
interpretation is that deterministic continuation makes clone trapping
easier to observe while added stochasticity weakens it. But the current
regime contrast is bundled: stochastic continuation changes both data
order and dropout relative to the deterministic continuation and
relative to the dropout-free parent. The paper should therefore present
the symmetry-breaking story as an interpretation of the outcome pattern,
not as a uniquely identified mechanism. The lag-budget split adds a
second qualification. Even within the deterministic regime, the reduced
seed-1 study shows that 16 steps are not enough to reveal the 128-step
reversal, and the direct mini-LAG baseline tables show the same failure
pattern. That qualification matters because it shows that ``a little
lag'' is helpful, but not automatically sufficient, for dense width
selection \citep{KarpEtAl2024}.

The newly exposed full-pool tables matter for the same reason. The
deterministic long-horizon reversal is not just \texttt{refsubspace}
versus one exact-copy baseline; it survives comparison against the
seed-0 asymmetric-reset families as well. That makes the deterministic
result harder to dismiss as a trivial optimizer-reset artifact, even
while the paper remains careful not to overstate it as a universally
better recipe.

\hypertarget{limitations}{%
\section{Limitations}\label{limitations}}

This study has clear limits.

\begin{itemize}
\tightlist
\item
  It is a single-dataset dense decoder-only proxy study
  \citep{EldanLi2023}.
\item
  The main evidence comes from one architecture family and a small
  number of checkpoints and seeds.
\item
  The stochastic regime is operational rather than canonical, because
  the parent was trained with zero dropout and stochasticity is
  introduced during continuation.
\item
  The deterministic-versus-stochastic comparison does not separately
  ablate shuffled order and dropout.
\item
  The non-clone challenger family is still narrow;
  \texttt{reference-subspace} is informative, but it is not the space of
  all meaningful asymmetric warm starts.
\item
  The deterministic 128-step reversals at step-2000 are small in
  magnitude even though they are consistent in sign.
\item
  The paper now exposes direct short-lag and full-pool evidence from
  existing artifacts, but it still lacks repeated continuation runs or
  uncertainty intervals.
\item
  The paper does not compare against frontier-scale pipelines and should
  not imply transfer to that setting
  \citep{DuEtAl2024, WortsmanEtAl2023}.
\end{itemize}

These limits are acceptable only because the claim is correspondingly
narrow.

\hypertarget{conclusion}{%
\section{Conclusion}\label{conclusion}}

Dense width growth in this study is best understood as a
candidate-selection problem over full training states whose answer
depends on regime and horizon. Zero-step preservation is not enough as a
universal selector. Exact-copy symmetric warm starts remain the
strongest short-budget and stochastic default, while a structured
non-clone challenger can win in longer deterministic continuation. The
most reliable low-cost selector is probe KL, not a universal escape
score. Under constrained compute, that is already a useful result: it
clarifies when preservation is enough, when it is not, and how much real
lag may be required before dense width selection becomes trustworthy.

\bibliographystyle{tmlr}
\bibliography{references}

\end{document}